\documentclass[preprint,12pt]{elsarticle}

\usepackage{amssymb}
\usepackage{amsmath}
\usepackage{amssymb}
\usepackage{xcolor}
\usepackage{booktabs}
\usepackage{multirow}
\usepackage{amsmath}
\usepackage{graphicx} 
\usepackage{colortbl}
\usepackage{wasysym}
\usepackage{array}
\usepackage{tabularx}
\usepackage{float}
\definecolor{bordercolor}{RGB}{215,215,215}
\definecolor{fillcolor}{RGB}{215,215,215}
\usepackage[T1]{fontenc}

\usepackage{graphicx,verbatim}
\usepackage{pifont}
\usepackage{booktabs}
\begin{document}
\makeatletter
\def\ps@pprintTitle{%
 \let\@oddhead\@empty
 \let\@evenhead\@empty
 \let\@oddfoot\@empty
 \let\@evenfoot\@empty
}
\makeatother

\begin{frontmatter}



\title{Enhancing Zero-Shot Brain Tumor Subtype Classification via Fine-Grained Patch-Text Alignment} 

\author[1,2]{Lubin Gan} 
\author[2]{Jing Zhang\corref{cor1}}
\author[3]{Linhao Qu} 
\author[1]{Yijun Wang} 
\author[2]{Siying Wu} 
\author[1,2]{Xiaoyan Sun\corref{cor1}}

\cortext[cor1]{Corresponding author} 

\affiliation[1]{organization={University of Science and Technology of China},
            city={Hefei},
            citysep={},
            postcode={230026}, 
            country={China}}

\affiliation[2]{organization={Anhui Province Key Laboratory of Biomedical Imaging and Intelligent Processing,Institute of Artificial Intelligence, Hefei Comprehensive National Science Center},
            city={Hefei},
            citysep={},
            postcode={230088}, 
            country={China}}

\affiliation[3]{organization={Fudan University},
            city={Shanghai},
            citysep={},
            postcode={200433},
            country={China}}

\begin{abstract}
The fine-grained classification of brain tumor subtypes from histopathological whole slide images is highly challenging due to subtle morphological variations and the scarcity of annotated data. Although vision-language models have enabled promising zero-shot classification, their ability to capture fine-grained pathological features remains limited, resulting in suboptimal subtype discrimination. To address these challenges, we propose the Fine-Grained Patch Alignment Network (FG-PAN), a novel zero-shot framework tailored for digital pathology. FG-PAN consists of two key modules: (1) a local feature refinement module that enhances patch-level visual features by modeling spatial relationships among representative patches, and (2) a fine-grained text description generation module that leverages large language models to produce pathology-aware, class-specific semantic prototypes. By aligning refined visual features with LLM-generated fine-grained descriptions, FG-PAN effectively increases class separability in both visual and semantic spaces. Extensive experiments on multiple public pathology datasets, including EBRAINS and TCGA, demonstrate that FG-PAN achieves state-of-the-art performance and robust generalization in zero-shot brain tumor subtype classification.
\end{abstract}

\begin{keyword}
Brain tumor classification \sep Zero-shot classification \sep Whole slide image \sep Vision-language model
\end{keyword}

\end{frontmatter}



\section{Introduction}

The classification of brain tumors based on whole slide images (WSIs) is a critical task in neuropathology, as it plays a crucial role in guiding therapeutic strategies and predicting patient outcomes~\cite{van2021deep,shao2021transmil,li2021dual,chen2024rethinking,glaser2017light,ludwig2005biomarkers,gong2022person,lin2025phys4dgen,gong2024beyond,gong2024beyondv2}. Traditional brain tumor diagnosis relies on manual microscopic examination, which is time-consuming and labor-intensive, significantly reducing diagnostic efficiency. With the rapid advancement of deep learning, an increasing number of researchers are focusing on automatically extracting features from high-resolution histopathological slices, enabling efficient and accurate classification of different types of brain tumors and achieving significant breakthroughs~\cite{webster2014whole,farahani2015whole,barker2016automated,ma2020brain,hou2016patch,elazab2024computer,mohan2024radpath,elazab2024multi}.

\begin{figure}[!t]
    \includegraphics[width=\textwidth]{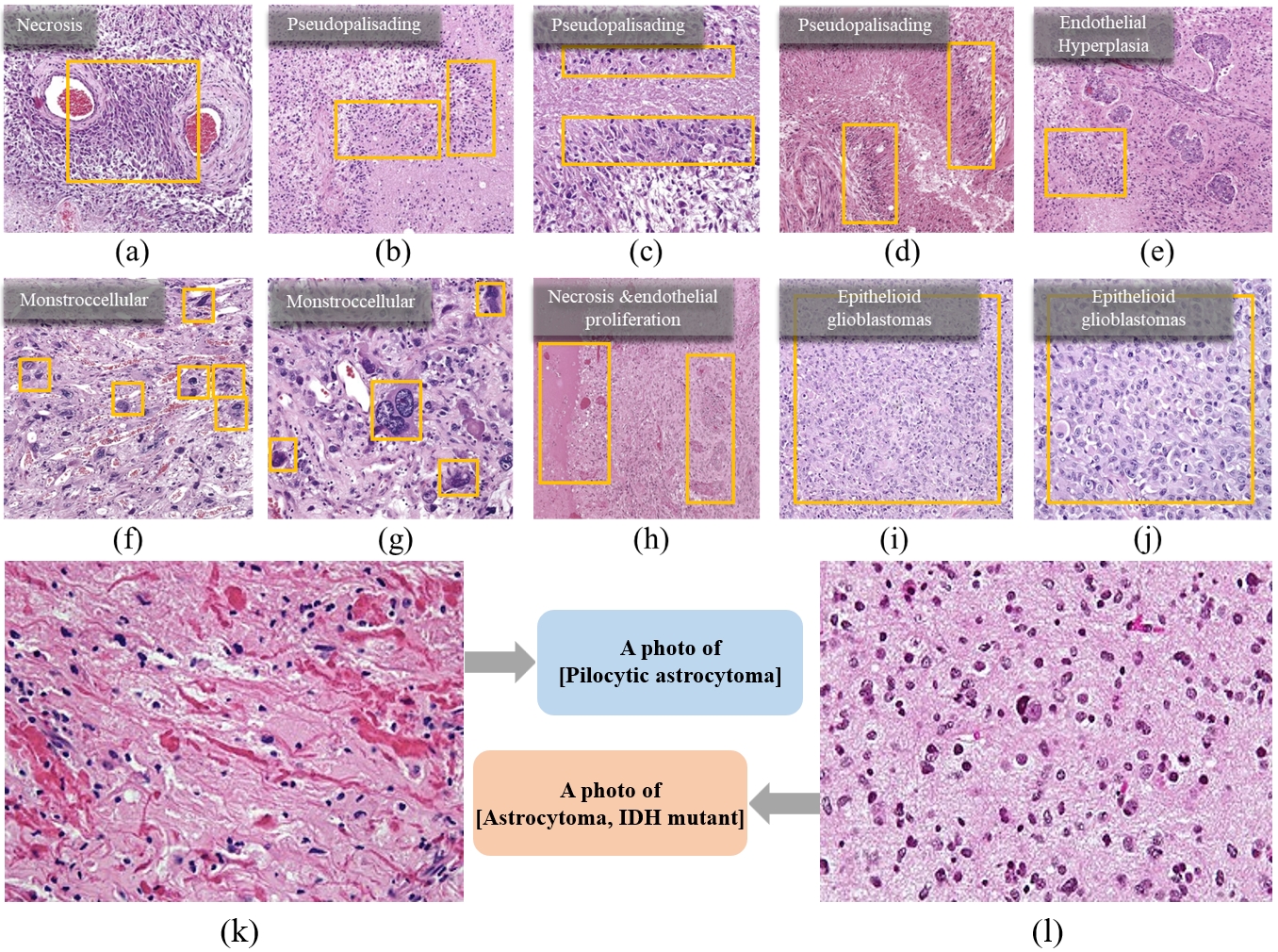}
    \caption{Illustration of representative histopathological patches with glioblastoma. (a) Viable tumor cells surround blood vessels within necrotic regions. (b) Serpentine-pattern necrosis with peripheral pseudopalisading of tumor cells. (c, d) High magnification of nuclear pseudopalisading around necrosis. (e) Endothelial proliferation with glomeruloid vessels and multilayering, alongside focal necrosis. (f, g) Presence of mono- or multinucleate giant cells with bizarre nuclei. (h) Mild to moderate nuclear pleomorphism, with both necrosis and endothelial proliferation supporting glioblastoma diagnosis. (i, j) Epithelioid glioblastoma subtype, featuring cohesive sheets of tumor cells with abundant cytoplasm and distinct borders. (k, l) Comparison between coarse-grained and fine-grained morphological features.
}
    \label{fig:patches}
\end{figure}

Although numerous supervised deep learning methods have been proposed for brain tumor classification~\cite{hsu2022weakly,li2023vision,ren2025triplane,ren2022dlformer,ren2021deep}, they rely heavily on costly and labor-intensive annotated datasets and often struggle to generalize effectively across new domains. Fine-grained pathological features, such as nuclear atypia, serpentine necrosis, and microvascular proliferation—provide crucial evidence for the discovery and accurate diagnosis of brain tumors, as illustrated in Figure \ref{fig:patches}. To address the dependence on large annotated datasets and improve generalizability, zero-shot classification methods~\cite{lu2023visual,javed2024cplip,alfasly2024zero,yin2025grpose,yin2025structure} have been introduced. For instance, TITAN~\cite{ding2024multimodal}, a pathology foundation model built on Vision Transformer, has demonstrated promising zero-shot performance. However, these foundation models still face notable limitations in capturing fine-grained pathological details~\cite{li2024generalizable,hu2024histopathology,zheng2024odtrack,zheng2023toward,zheng2022leveraging,zheng2025decoupled,zheng2025towards}, hindering their ability to achieve high performance on challenging tasks such as fine-grained brain tumor classification.

To address the challenges of high morphological similarity among different brain tumor subtypes and enhance their zero-shot classification performance, we have developed a method called the Fine-Grained Patch Alignment Network (FG-PAN). The core of this method is to synthesize more discriminative local features for each patch and class through an integrated strategy, aiming to significantly widen the gap between different class representations in the feature space. To this end, we introduce a local feature refinement module, which enhances visual features by modeling the spatial relationships between patches. Concurrently, a fine-grained text description generation module leverages large language models (LLMs) to generate richer pathological descriptions for each category, thereby effectively increasing the distance between different classes in the semantic space. Final slide-level predictions are then obtained through an uncertainty-aware aggregation strategy. We argue that precise classification at the patch level is a critical yet underexplored direction within the context of brain tumor subtype classification.

In summary, our main contributions are as follows:
\begin{itemize}
    \item We propose FG-PAN, a novel fine-grained patch alignment network that synthesizes more discriminative local features for each patch and class through an integrated strategy, significantly improving the separability of class representations in the feature space for zero-shot brain tumor classification; 
    \item We introduce a local feature refinement module and a fine-grained text description generation module, which leverage spatial relationships among patches and LLM-generated pathology-aware descriptions, respectively, to enhance both visual and semantic discriminability; 
    \item Extensive experiments on multiple digital pathology datasets demonstrate that FG-PAN achieves state-of-the-art performance on brain tumor classification benchmarks and exhibits strong generalization to out-of-domain datasets, highlighting its robustness and transferability in computational pathology. 
\end{itemize}

\section{Related Work}

In this section, we review the relevant researches and discuss recent advancements, including pathology classification, zero-shot classification using VLMs, prompt engineering.

\subsection{Pathology classification}

Histopathological image classification stands as a central task in digital pathology, aiming to facilitate tumor subtype identification, prognosis prediction, and molecular profiling through the analysis of high-resolution WSIs~\cite{qu2022towards}. Recent advances in this domain predominantly fall into two methodological paradigms: Pathology Foundation Models (PFMs)~\cite{chen2024towards,zhou2024knowledge,vorontsov2024foundation,wang2023brightness,peng2021ensemble,ren2024ultrapixel,yan2025textual,peng2024efficient,conde2024real,peng2025directing,peng2025pixel,peng2025boosting,he2024latent,di2025qmambabsr,peng2024unveiling,he2024dual,he2024multi,pan2025enhance} and conventional supervised learning approaches~\cite{qu2022bi,qu2022dgmil,qu2024rethinking,qu2023rise,qu2023boosting,fei2025whole,sezer2020convolutional,xu2016deep,he2023predicting,yu2021convolutional,chen2021automatic,li2023ntire,ren2024ninth,wang2025ntire,peng2020cumulative,wang2023decoupling,peng2024lightweight,peng2024towards,wu2025dropout}. While conventional supervised methods have long served as the dominant paradigm, recent years have witnessed a growing interest in PFMs~\cite{xiong2025survey,lin2025jarvisart,lin2025jarvisir,lin2024aglldiff,lin2024unsupervised,lin2024fusion2void,wang2023learning,lin2023domain,wang2025learning,he2024diffusion,he2025segment,he2023hqg,he2025unfoldir,he2025run,he2025reti,he2024weakly,xiao2024survey,he2023strategic,he2023camouflaged,he2023degradation}. PFMs represent a novel class of models proposed in recent years, leveraging large-scale pretraining on extensive pathology datasets to acquire transferable and generalizable representations. These representations have demonstrated strong performance across a wide range of downstream histopathological classification tasks.

Initial developments in PFMs, exemplified by CTransPath~\cite{wang2022transformer}, primarily concentrated on optimizing patch-level feature extractors via contrastive learning objectives. Subsequent advancements, such as HIPT~\cite{chen2022scaling}, UNI~\cite{chen2024towards}, and Virchow~\cite{vorontsov2024foundation}, have incorporated state-of-the-art self-supervised learning paradigms—most notably DINO~\cite{caron2021emerging}, MAE~\cite{he2022masked} and DINOv2~\cite{oquab2023dinov2}) to enable representation learning at unprecedented scale, encompassing billions of image patches. In parallel, a growing body of work has shifted toward aggregator-centric PFMs, including CHIEF~\cite{wang2024pathology}, TITAN~\cite{ding2024multimodal}, THREADS~\cite{vaidya2025molecular}, which enhance slide-level modeling capacity through the integration of multimodal pretraining strategies and vision-language alignment. These models not only improve global contextual reasoning but also offer substantial reductions in computational cost, paving the way for more scalable and semantically enriched whole-slide representations.

Conventional supervised approaches have served as a foundation for pathological image classification, typically leveraging annotated datasets to train CNNs or ViT for feature extraction. Despite their effectiveness, these methods suffer from key limitations: they rely heavily on expert-labeled data, which is costly and labor-intensive to obtain; they exhibit poor generalization across domains such as different organs or staining protocols; and they often lack the granularity needed to distinguish morphologically similar subtypes. These challenges restrict their scalability and adaptability in real-world pathological scenarios.

While PFMs offer greater scalability and generalizability, they often fall short in capturing subtle morphological cues essential for fine-grained classification.  In particular, existing models struggle to identify fine-grained features such as nuclear atypia, cytoplasmic texture, and glandular structures.  Furthermore, issues related to end-to-end pretraining, aggregator scalability, and domain adaptability remain open.  In contrast, supervised methods—though less scalable—tend to excel at detailed morphological discrimination due to task-specific optimization.  However, their strong reliance on expert annotations and poor cross-domain generalization significantly limit their broader applicability.

\subsection{Zero-shot classification using VLMs}

Vision-language models (VLMs)~\cite{furst2022cloob,jia2021scaling,radford2021learning,singh2022flava,yuan2021florence,jiang2024dalpsr,ignatov2025rgb,du2024fc3dnet,jin2024mipi,sun2024beyond,qi2025data,feng2025pmq,xia2024s3mamba,pengboosting,suntext} are designed to learn joint representations of visual and textual modalities, enabling effective cross-modal alignment between images and their associated textual descriptions. By constructing a shared embedding space, these models demonstrate strong zero-shot classification performance, particularly when queried with carefully designed textual prompts such as class names during inference.

Among them, FLAVA~\cite{singh2022flava} incorporates both aligned and unaligned image-text inputs and optimizes for multimodal as well as unimodal objectives to enhance general cross-modal understanding. ALIGN~\cite{jia2021scaling} leverages a massive corpus of loosely aligned image-alt-text pairs, optimizing a contrastive loss to learn visual-semantic associations from large-scale, noisy web data. In contrast, CLIP~\cite{radford2021learning} focuses on training over a relatively cleaner but smaller dataset of curated image-text pairs, employing a dual-encoder architecture to maximize agreement between matched pairs and suppress irrelevant ones using a contrastive learning framework.

At inference time, CLIP computes the similarity between an input image and a set of textual class prompts, often in the form of natural language templates like “a photo of a [class name]”, and assigns the image to the class with the highest similarity score. Notably, CLIP’s performance can be further improved via prompt engineering, where manually crafted prompts lead to better alignment and classification accuracy in zero-shot scenarios.

\begin{figure}[t]
    \includegraphics[width=\textwidth]{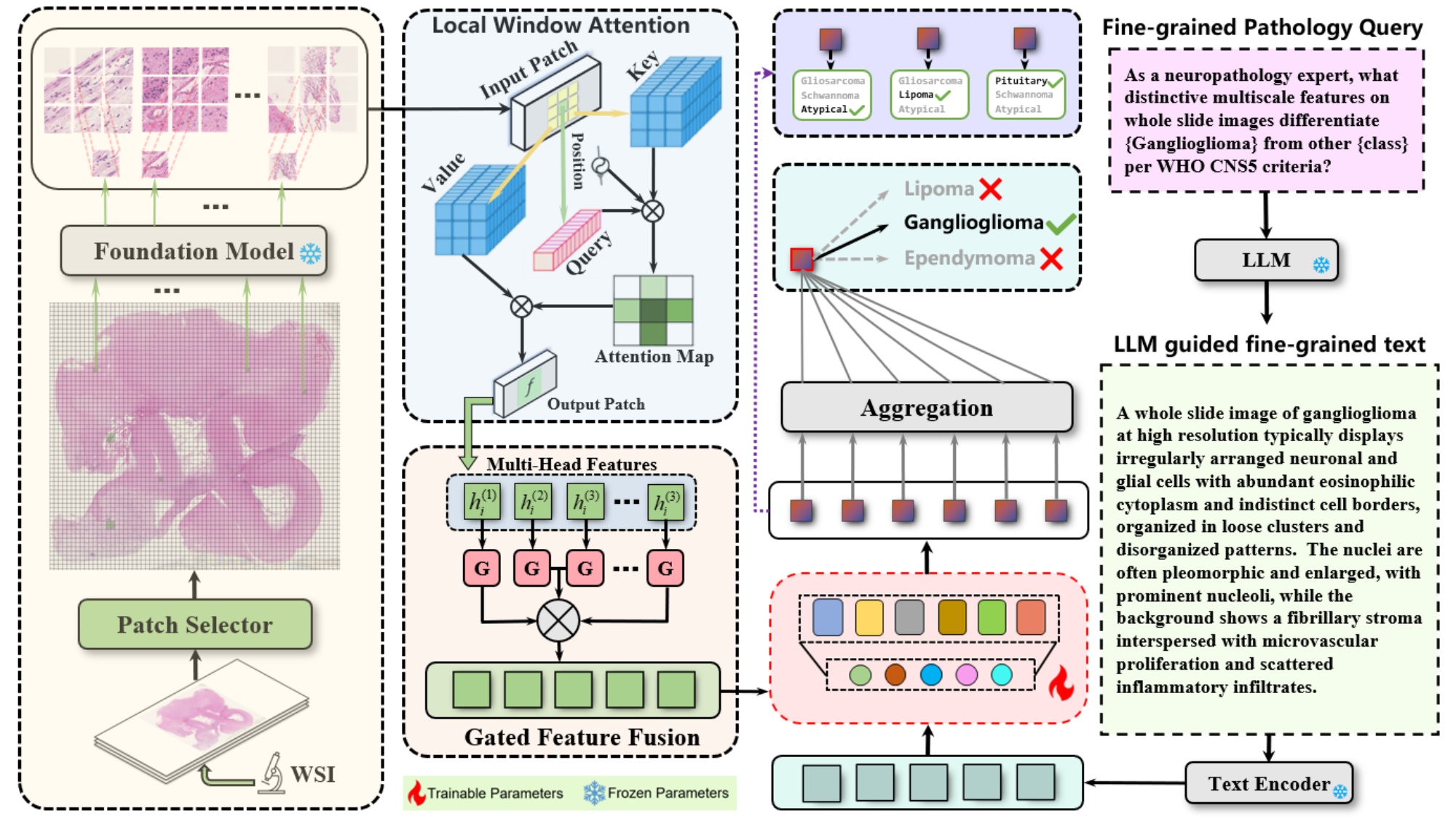}
    \caption{Overview of the FG-PAN framework. In the visual flow, FG-PAN first selects representative patches from the WSI, then refines their visual features via local window attention and gated fusion modules. In the textual flow, FG-PAN prompts pathological LLM to generate class-specific fine-grained textual descriptions for constructing semantic prototypes. Patch-level visual and semantic features are aligned for zero-shot classification, and a coordinate-aware aggregation mechanism is introduced to predict the slide-level results.}
    \label{fig:mainframe}
\end{figure}

\subsection{Prompt engineering}

Recent advances in prompt tuning~\cite{gan2023decorate,jia2022visual,khattak2023maple,shu2022test,zhu2023prompt,lu2024mace,lu2023tf,lu2024robust,li2025set,gao2024eraseanything,duan2025dit4sr,duan2025diffretouch,AAAI_2025_DiffOOM} have largely focused on optimizing class-level text embeddings to improve zero-shot performance in vision-language models. CoOp~\cite{zhou2022learning} enhances classification by appending learnable context vectors to textual prompts, while CoCoOp~\cite{zhou2022conditional} further conditions prompt learning on the image itself. However, such methods typically require fine-tuning and demonstrate limited generalization to unseen or fine-grained categories, often falling short of CLIP's zero-shot baseline when applied to novel domains.

An alternative paradigm explores using large language models (LLMs) to generate enriched prompts or attribute descriptions.  Works such as CHiLS~\cite{novack2023chils} utilize GPT-based taxonomic refinements for hierarchical classification, while CuPL~\cite{pratt2023does} and Menon et al.~\cite{menon2022visual} integrate GPT-derived class attributes into hand-crafted prompt templates (e.g., “a photo of a [class]”) to improve general-domain performance.  Despite their success, these approaches often underperform in fine-grained domains like computational pathology, where subtle morphological cues—such as nuclear atypia, cytoplasmic texture, or glandular architecture—are essential for accurate subtype differentiation.

In this work, we propose a plug-and-play strategy tailored for digital pathology, where textual prompts are enriched using LLM-generated descriptions that explicitly encode histopathological attributes of each tumor subtype.  Without modifying any parameters of the vision-language model, our approach provides refined prompts that align more effectively with patch-level morphological features commonly observed in WSIs. This enables more accurate zero-shot classification in challenging fine-grained pathology tasks, while maintaining full compatibility with pretrained VLMs like CLIP.

\section{Problem Setup}

\subsection{Problem Definition}
In general pathological classification, the fine-grained brain tumor subtype classification task involves dividing a high-resolution WSI $I$ into $M$ patches ${\{x_i\}^M_{i=1}}$, where each patch $x_i$ represents a fixed-size subregion of the WSI. Each WSI is associated with a global label $y\in C$, and the goal is to predict $y$ using local features from $x_i$ and its surrounding patches $w(x_i)$. Specifically, features $f_i$ are extracted from $x_i$ and $w(x_i)$ using a feature extractor $\varepsilon$:
\begin{equation}
{f_i} = \varepsilon ({x_i},w({x_i})),
\end{equation}
and local predictions $\hat{y_i}$ are generated for each patch. These predictions are aggregated using a function $\varsigma$, which combines spatial positions and visual features to produce the global prediction $\hat{y_i}$:
\begin{equation}
\hat{y}  = \varsigma (\{ {\hat{y_i}},({x_i},{y_i})\} {}_{i = 1}^M).
\end{equation}
The loss function $\Gamma$ balances patch-level and slide-level classification losses to ensure both local and global consistency.

\subsection{Motivation}

\begin{figure}[t]
    \includegraphics[width=\textwidth]{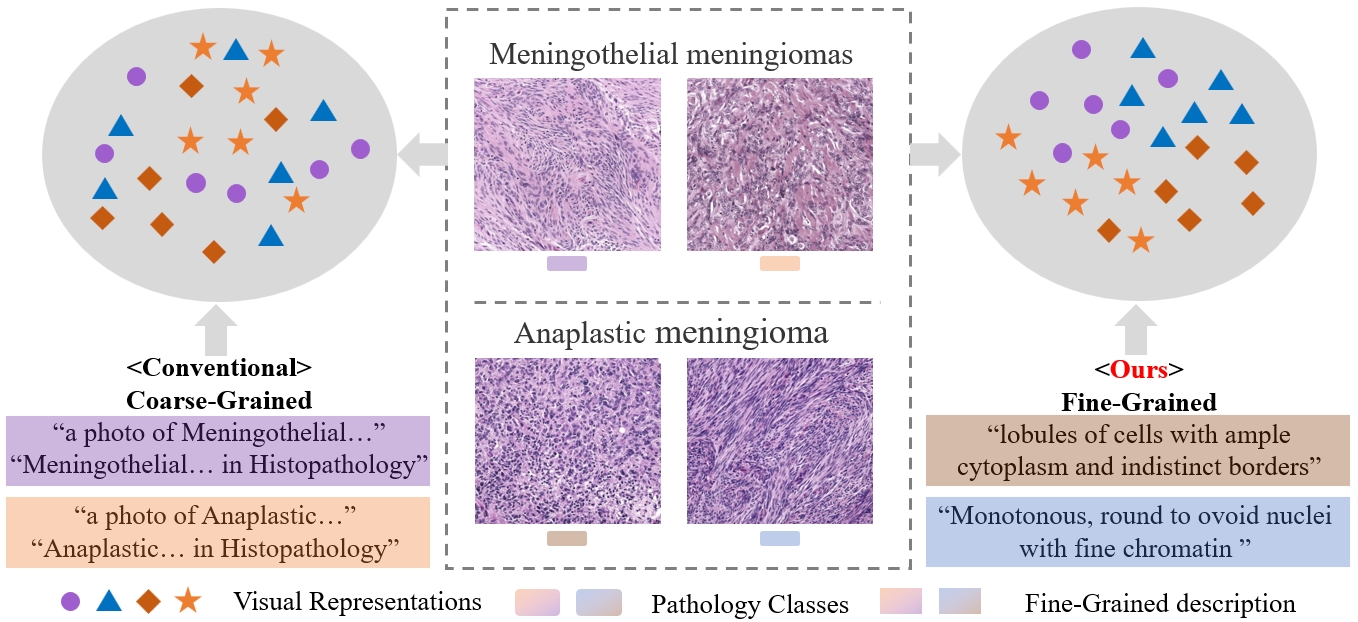}
    \caption{The left panel illustrates conventional, coarse-grained text prompts, which provide limited semantic detail and result in less discriminative visual-semantic alignment. The right panel demonstrates the proposed fine-grained approach, where pathology-aware, LLM-generated descriptions provide richer semantic information and enable more precise alignment between visual features and class prototypes.  The schematic highlights improved class separability and semantic interpretability in the feature space when using fine-grained prompts.}
    \label{fig:compare}
\end{figure}

Recent advances in VLMs have enabled zero-shot classification by aligning visual features with textual semantics.      However, applying such models to pathological images remains challenging due to several inherent limitations.      Existing approaches often \textbf{struggle to capture the fine-grained morphological variations} that distinguish pathological subtypes.  Generic textual prompts typically \textbf{lack the semantic specificity} required to represent subtle visual cues, such as nuclear atypia, glandular architecture, or cytoplasmic texture,which are critical for clinical diagnosis.   As illustrated in Figure \ref{fig:compare}, conventional coarse-grained prompts provide only limited guidance for visual-semantic alignment, resulting in less discriminative feature representations and overlapping class boundaries. In contrast, fine-grained, pathology-aware prompts offer richer semantic information, leading to clearer separation between subtypes in the feature space.   Moreover, many methods process image patches independently and fail to model the \textbf{spatial dependencies across patches}, resulting in fragmented contextual understanding.      Additionally, models trained on specific datasets frequently exhibit poor generalization when applied to external cohorts, reflecting their limited robustness across tissue types, staining protocols, and imaging domains.

To address these challenges, we propose FG-PAN, a unified framework designed to enhance zero-shot classification in digital pathology by incorporating both semantic prompting and spatial awareness.      Specifically, we introduce \textbf{a structure-aware prompt generation module} that embeds morphological attributes into class descriptions, enabling more precise alignment between patch-level visual features and their corresponding textual semantics.      Furthermore, we develop \textbf{a coordinate-aware aggregation module} that explicitly encodes spatial relationships among patches, thereby facilitating context-aware reasoning across the WSI.      These components are seamlessly integrated into a prompt-adaptive classification pipeline, which operates \textbf{without requiring additional supervision}, while improving both robustness and generalization.      Extensive experiments on multiple public pathology datasets demonstrate that FG-PAN consistently outperforms strong baselines, validating the effectiveness and broad applicability of our method.

\section{Method}

In this section, we first introduce the problem formulation of our zero-shot classification task. Then, we give a detailed description of the proposed critical components in the following subsections.

\subsection{Framework Overview}
The overall framework of FG-PAN is illustrated in Fig \ref{fig:mainframe}. This framework is designed to address the fine-grained zero-shot classification problem of brain tumor subtypes by explicitly aligning localized histopathological features with discriminative textual attributes. The pipeline begins with preprocessing WSIs, where the regions containing tissues are segmented and divided into fixed-size patches. However, in digital pathology, the majority of patches may contain non-informative background or benign tissue and are not diagnostically relevant. To address this, we introduce a patch selection module to filter out irrelevant regions and identify a small subset of representative patches. These patches contain discriminative morphological features for subtype classification.

Each selected patch is encoded using a frozen pathology foundation model, producing visual embeddings that capture localized cellular and structural characteristics. Then, we design a feature refinement module to model spatial interactions among neighboring patches for refining patch embeddings. Specifically, the local window attention mechanism is first utilized to aggregate context-aware cues within each patch’s spatial neighborhood. Following this, a gated fusion unit is combined to integrate the multi-head outputs, strategically suppressing irrelevant signals while amplifying diagnostically salient features. The refined patch representations are then used to compute patch-level classification scores by comparing them with the textual prototypes.

On the other hand, we devise the fine-grained text description generation module to generate the textual prototypes, which serve as semantically rich and pathology-aware class descriptions. The prototypes are guided to incorporate both histological patterns and molecular attributes, enabling a fine-grained alignment with visual morphological features. This module can refresh model generalization to unseen categories in the zero-shot setting.

Finally, FG-PAN aggregates all patch-level predictions using a coordinate-aware weighting mechanism that adaptively fuses scores from the most representative patches based on their spatial positions. The final slide-level classification results are obtained.

\subsection{Local Window Attention}

To enhance the modeling of spatial dependencies between histopathological patches, we design a Local Window Attention module that focuses on context-aware feature refinement within localized windows. Compared to global attention, this localized mechanism better captures fine-grained patterns crucial for distinguishing tumor subtypes with subtle morphological differences. 

To begin, the WSI will be divided into $\mathit{M}$ fixed-size patches$\{ {x_i}\} _{i = 1}^M$, each with a corresponding embedding ${f_i} \in {\mathbb{R}^d}$ and spatial coordinate $({x_i},{y_i})$. The input patch features are first partitioned into $\mathit{N}$ equalsized, non-overlapping local windows $\{ {W_k}\} _{k = 1}^N$, each containing $S \times S$ patches based on their spatial proximity.

For each window $\mathit{W_k}$, we concatenate the features of the patches it contains to form a local feature matrix ${F_k} \in {\mathbb{R}^{{S^2} \times d}}$. This matrix is passed through a multi-head self-attention block to model intra-window dependencies: 
\begin{align}
Q_k &= F_k W_Q, \quad K_k = F_k W_K, \quad V_k = F_k W_V \\
A_k &= \text{Softmax}\left( (Q_k K_k^\top + B) / \sqrt{d} \right), \quad O_k = A_k V_k,
\end{align}
where ${W_Q},{W_K},{W_V} \in {\mathbb{R}^{d \times d}}$ are learnable projection matrices, $B \in {\mathbb{R}^{{S^2} \times {S^2}}}$ is a learnable relative positional bias matrix, encoding the spatial layout within each window, ${A_k}$ is the attention score matrix, and ${O_k} \in {\mathbb{R}^{{S^2} \times d}}$ represents the refined patch features in window ${W_k}$.

\subsection{Gated Feature Fusion}

In histopathology analysis, individual tissue blocks often exhibit high within-group variance due to sampling heterogeneity, while multiple tissue blocks across WSIs may redundantly encode similar structures. To obtain a robust and discriminative representation for each patch, it is necessary to selectively emphasize informative features and suppress irrelevant or noisy cues. To this end, we employ a gated feature fusion mechanism, which is a lightweight, learnable attention-like module applied to multi-head local outputs. The mechanism operates independently on each patch and integrates its multi-source features through dynamic gating. Specifically, it selectively aggregates features obtained from the multi-head outputs of the previous Local Window Attention module.

Let $\{ h_i^{(1)},h_i^{(2)}, \ldots ,h_i^{(L)}\}  \in {\mathbb{R}^d}$ denote the attention-refined representations of patch $i$ from $L$ heads. We compute a learned importance weight $\lambda _i^{(l)} \in \left[ {0,1} \right]$ for each feature via:
\begin{equation}
\gamma_i^{(l)} = \sigma\left(W_g^{(l)} h_i^{(l)} + b_g^{(l)}\right), \quad l = 1, \ldots, L,
\end{equation}
where $W_g^{(l)} \in {\mathbb{R}^{d \times 1}}$ and $b_g^{(l)} \in {\mathbb{R}}$ are the head-specific gating parameters, and $\sigma ( \cdot )$ is a sigmoid function. Inspired by the gating units in the attention mechanism, we find it is particularly suitable for the patch-level aggregation problem, which is particularly important for WSI analysis where the spatial patterns are complex and fragmented.

Next, the head-specific features are modulated by their respective gating weights:
\begin{equation}
\hat{h}_i^{(l)} = \gamma_i^{(l)} \odot h_i^{(l)}.
\end{equation}
This element-wise multiplication yields filtered feature vectors that retain only semantically meaningful dimensions. The fused representation is then computed by linearly projecting the sum of gated features:
\begin{equation}
H_i = W_f \left( \sum_{l=1}^{L} \hat{h}_i^{(l)} \right) + b_f,
\end{equation}
where $W_f \in {\mathbb{R}^{d \times d}}$ and $b_f \in {\mathbb{R}^d}$ are the learnable projection parameters.

\subsection{Fine-grained Text Description Generation}

Vision-language models (VLMs) have shown impressive generalization capabilities in natural image domains by learning a shared embedding space between visual inputs and their textual descriptions.  However, directly applying such models to the domain of histopathology, especially for fine-grained tumor subtype classification—poses significant challenges.  Unlike natural categories that can be described by common language (e.g., “a photo of a dog”), pathological subtypes require nuanced semantic distinctions rooted in expert medical knowledge.

To address this gap, we propose the Fine-grained Text Description Generation, which leverages the capabilities of LLMs to generate structured, fine-grained class descriptions.  These serve as textual prototypes in the zero-shot classification framework, enabling cross-modal alignment with patch-level visual features extracted from WSIs.

Rather than relying on simple category names, we design diagnostically enriched prompts to query the LLM with domain-specific context. For each tumor subtype $c$, we query the LLM with prompts such as:
\begin{center}
    \fcolorbox{bordercolor}{fillcolor}{%
        \parbox{0.95\textwidth}{%
            "As a neuropathology expert, what distinctive multiscale features on whole slide images differentiate [class] from other [class] per WHO CNS5 criteria? Generate discriminative attribute pairs combining molecular profiles and histopathological signatures using the format: '[class] with [molecular feature] and [histological pattern]'."
        }
    }
\end{center}
[class]: Refers to a specific category or subclass. In fine-grained classification tasks, [class]: WHO-defined specific tumor entity (e.g., "Anaplastic oligodendroglioma, IDH-mutant and 1p/19q-codeleted"), [molecular feature]: Essential WHO-mandated alterations (e.g., "IDH1 R132H mutation with 1p/19q whole-arm codeletion"), [histological pattern]: Diagnostic structural signatures (e.g., "microcystic honeycomb architecture containing fried-egg cells").

It is worth highlighting that this is a plug-and-play module, requiring no retraining or fine-tuning, thus offering good practicality and efficiency. As shown in Fig. \ref{fig:tsne}, improved prompts demonstrate enhanced interclass separability in the feature space through text reconstruction, significantly optimizing the discriminative potential of pathological semantic embedding vectors, thus improving classification performance.

\subsection{Patch-wise Cross-model Classification}

In fine-grained zero-shot classification, it is essential to compare visual instances (i.e., patch features) against semantically rich class descriptions in a shared embedding space. To this end, we formulate patch-level cross-modal classification as a similarity-based retrieval problem, where each patch embedding is matched against a set of textual class prototypes.
Based on this, let $H_i \in \mathbb{R}^d$ denote the refined visual feature of patch $i$, as produced by the previous gated fusion module. Let $\{ {T_c}\} _{c = 1}^C$ denote the set of textual class embeddings, where each $T_c \in \mathbb{R}^d$ is a semantic prototype generated from pathology-informed textual descriptions (see Sec. 4.4).  These textual embeddings are normalized and serve as reference anchors in the joint space. We compute the similarity between each patch and class prototype using the cosine similarity function, scaled by a learnable temperature parameter $\tau  > 0$:
\begin{equation}
s_i^{(c)} = \frac{H_i^\top T_c}{\|H_i\| \cdot \|T_c\|}, \quad
p_i^{(c)} = \frac{\exp(s_i^{(c)} / \tau)}{\sum_{k=1}^{C} \exp(s_i^{(k)} / \tau)},
\end{equation}
where $s_i^{(c)}$ denotes the raw cosine similarity score between patch $i$ and class $c$, $p_i^{(c)}$ represents the softmax-normalized probability over classes, $\tau$ controls the sharpness of the distribution and is jointly optimized during training.

For training, we adopt a standard cross-entropy loss over the predicted patch probabilities. Given a training patch $x_i$ with label $y_i$, the classification loss is defined as:
\begin{equation}
\mathcal{L}_{\text{patch}} = -\log\left(p_i^{(y_i)}\right).
\end{equation}

\subsection{Coordinate-aware Aggregation}

While each patch in WSI is independently classified based on its visual-textual alignment, these predictions must be aggregated to produce a consistent pathology diagnosis. Existing methods ignore the anatomical and spatial organization of histological features, which are often key to fine-grained pathology classification. Therefore, we design a coordinate-aware aggregation module that fuses patch predictions based on semantic confidence and spatial location.

We compute a learnable importance weight ${\alpha _i} \in \left[ {0,1} \right]$ for each patch, which reflects both its visual semantics and positional context. Specifically, we first encode the spatial location using a trainable positional embedding function $\phi ({x_i},{y_i}) \in \mathbb{R}^d$, and concatenate it with the patch’s final visual representation $H_i$. The importance is computed via:
\begin{equation}
\alpha_i = \frac{\exp\left(w^\top [H_i \| \phi(x_i, y_i)]\right)}{\sum_{j=1}^{N} \exp\left(w^\top [H_j \| \phi(x_j, y_j)]\right)},
\end{equation}
where $w \in \mathbb{R}^{2d}$ is a learnable projection vector, $||$ denotes feature concatenation, $\phi((x_i, y_i))$ can be implemented as a sinusoidal encoding, learned embedding table. The final WSI-level prediction is computed as a weighted sum of patch-level distributions:
\begin{equation}
P(y \mid \mathrm{WSI}) = \sum_{i=1}^{N} \alpha_i \cdot p_i.
\end{equation}


\section{Experimental Results and Discussions}

\subsection{Datasets}
For training, we use the \textbf{CAMELYON}~\cite{litjens20181399} dataset, which includes the WSIs of sentinel lymph node tissue sections. 500 WSIs diagnosed with lung cancer are chosen from the CAMELYON repository. We segment each WSI into 200~300 representative patches. For testing, we adapt disease categories that are purely different from training data to ensure zero-shot settings, including four test datasets, which are described as below.

\textbf{EBRAINS}. The Digital Tumor Atlas (EBRAINS)~\cite{roetzer2022digital} is a histopathological dataset developed by the University of Vienna, containing high-resolution H\&E-stained FFPE whole slide images (WSIs) of central nervous system tumors. It covers 32 histological subtypes, with expert-verified labels. After excluding 2 classes with fewer than 5 samples, 30 fine-grained subtypes were retained and grouped into 12 coarse-grained categories based on histological similarity. A total of 1200 WSIs were selected using stratified random sampling with a fixed seed, preserving class distributions at both granularity levels.

\textbf{IPD-Brain}. The IPD-Brain~\cite{chauhan2024ipd} is a histopathological dataset collected from the Nizam’s Institute of Medical Sciences (NIMS) in Hyderabad, India. It includes 547 H\&E-stained WSIs scanned at 40× magnification from 367 patients. Among these, 484 slides with complete clinical annotations correspond to three glioma subtypes: Glioblastoma, Astrocytoma, and Oligodendroglioma. Each slide includes metadata such as WHO grade and IHC biomarker status. All WSIs are derived from diagnostic specimens collected under clinical protocols.

\textbf{TCGA}. The Cancer Genome Atlas (TCGA)~\cite{tomczak2015review} is a publicly funded project that aims to catalog and discover major cancer-causing genomic alterations to create a comprehensive "atlas" of cancer genomic profiles. Our TCGA dataset specifically includes two types of tumors: Glioblastoma and Low-Grade Glioma. For evaluation, the dataset comprises 262 WSIs for GBM and 167 WSIs for LGG, totaling 429 WSIs. These samples were randomly selected from the TCGA brain tumor database, ensuring a representative and diverse set of cases for our experiments. We conducted a binary classification task on this dataset.

\textbf{BRACS}. The BRACS~\cite{brancati2022bracs} is a histopathological dataset specifically designed for breast carcinoma subtyping, developed through a collaboration between the National Cancer Institute—IRCCS “Fondazione Pascale,” the Institute for High Performance Computing and Networking (ICAR) of the National Research Council of Italy, and IBM Research-Zurich. The dataset comprises 547 H\&E-stained WSIs and 4539 Regions of Interest (ROIs) extracted from WSIs, all annotated by consensus among three board-certified pathologists. It covers three main lesion types (Benign, Atypical, and Malignant), further subdivided into seven subtypes.  Notably, atypical lesions (FEA and ADH) are included as precancerous subtypes, offering unique opportunities for early diagnosis research.

\begin{table}[t]
    \centering
    \renewcommand\arraystretch{1}
    \setlength{\tabcolsep}{3pt} 
    {
        \tiny
        \begin{tabular}{@{}ll*{14}{c}@{}}
            \toprule
            \multirow{2}{*}{Dataset} & \multirow{2}{*}{Metric} & \multicolumn{2}{c}{UNI~\cite{chen2024towards}} & \multicolumn{2}{c}{Conch~\cite{lu2024visual}} & \multicolumn{2}{c}{Conch-v1.5~\cite{ding2024multimodal}} & \multicolumn{2}{c}{Virchow~\cite{vorontsov2024foundation}} & \multicolumn{2}{c}{PLIP~\cite{huang2023visual}} & \multicolumn{2}{c}{CTransPath~\cite{wang2022transformer}}  \\
            \cmidrule(lr){3-4} \cmidrule(lr){5-6} \cmidrule(lr){7-8}
            \cmidrule(lr){9-10} \cmidrule(lr){11-12} \cmidrule(lr){13-14}
            & & Base & Ours & Base & Ours & Base & Ours & Base & Ours & Base & Ours & Base & Ours \\ 
            \midrule
            \multirow{3}{*}{EBRAINS$_{\text{30}}$} 
              & Bal. acc. & 0.212 & \bfseries 0.228 & 0.357 & \bfseries 0.405 & 0.493 & \bfseries 0.572 & 0.412 & \bfseries 0.465 & 0.247 & \bfseries 0.285 & 0.205 & \bfseries 0.224 \\
              & F1         & 0.241 & \bfseries 0.257 & 0.365 & \bfseries 0.427 & 0.542 & \bfseries 0.597 & 0.425 & \bfseries 0.474 & 0.358 & \bfseries 0.362 & 0.231 & \bfseries 0.239 \\
              & AUROC      & 0.853 & \bfseries 0.861 & 0.828 & \bfseries 0.859 & 0.913 & \bfseries 0.923 & 0.870 & \bfseries 0.895 & 0.883 & \bfseries 0.905 & \bfseries 0.895 & 0.856 \\
            \midrule
            
            \multirow{3}{*}{EBRAINS$_{\text{12}}$}
              & Bal. acc. & 0.259 & \bfseries 0.289 & 0.386 & \bfseries 0.396 & 0.635 & \bfseries 0.712 & 0.489 & \bfseries 0.522 & 0.322 & \bfseries 0.345 & 0.227 & \bfseries 0.249 \\
              & F1         & 0.269 & \bfseries 0.298 & 0.391 & \bfseries 0.397 & 0.657 & \bfseries 0.737  & 0.495 & \bfseries 0.526 & 0.415 & \bfseries 0.442 & 0.236 & \bfseries 0.258 \\
              & AUROC      & 0.864 & \bfseries 0.895 & \bfseries 0.882 & 0.865 & 0.927 & \bfseries 0.938  & 0.892 & \bfseries 0.903 & 0.905 & \bfseries 0.908 & 0.852 & \bfseries 0.861 \\
            \midrule

            \multirow{3}{*}{IPD-Brain$_{\text{3}}$}
              & Bal. acc. & 0.325 & \bfseries 0.368 & 0.347 & \bfseries 0.382 & 0.486 & \bfseries 0.522  & 0.382 & \bfseries 0.390 & 0.277 & \bfseries 0.310 & \bfseries 0.305 & 0.282 \\
              & F1         & 0.409 & \bfseries 0.493 & 0.408 & \bfseries 0.515 & 0.412 & \bfseries 0.530  & 0.410 & \bfseries 0.455 & 0.440 & \bfseries 0.470 & \bfseries 0.455 & 0.443 \\
              & AUROC      & 0.787 & \bfseries 0.803 & 0.786 & \bfseries 0.816 & 0.791 & \bfseries 0.833  & 0.792 & \bfseries 0.825 & 0.811 & \bfseries 0.843 & \bfseries 0.825 & 0.816 \\
            \midrule
            
            \multirow{3}{*}{BRACS$_{\text{7}}$}
              & Bal. acc. & 0.251 & \bfseries 0.287 & 0.253 & \bfseries 0.298 & 0.289 & \bfseries 0.329  & 0.265 & \bfseries 0.298 & 0.231 & \bfseries 0.260 & 0.218 & \bfseries 0.236 \\
              & F1         & 0.243 & \bfseries 0.282 & 0.236 & \bfseries 0.270 & 0.291 & \bfseries 0.334  & 0.251 & \bfseries 0.285 & 0.268 & \bfseries 0.295 & 0.285 & \bfseries 0.293 \\
              & AUROC      & 0.768 & \bfseries 0.771 & 0.772 & \bfseries 0.793 & 0.786 & \bfseries 0.796  & 0.774 & \bfseries 0.796 & 0.783 & \bfseries 0.810 & \bfseries 0.796 & 0.793 \\
            \midrule
            
            \multirow{3}{*}{TCGA$_{\text{2}}$}
              & Bal. acc. & 0.712 & \bfseries 0.753 & 0.763 & \bfseries 0.785 & 0.679 & \bfseries 0.797 & 0.735 & \bfseries 0.772 & 0.713 & \bfseries 0.745 & 0.710 & \bfseries 0.731 \\
              & F1         & 0.736 & \bfseries 0.789 & 0.779 & \bfseries 0.802 & 0.712 & \bfseries 0.801 & 0.750 & \bfseries 0.782 & 0.710 & \bfseries 0.769 & 0.782 & \bfseries 0.802 \\
              & AUROC      & 0.910 & \bfseries 0.926 & \bfseries 0.932 & 0.910 & 0.772 & \bfseries 0.891 & 0.923 & \bfseries 0.941 & 0.891 & \bfseries 0.928 & 0.902 & \bfseries 0.910 \\
            \bottomrule
        \end{tabular}
    }
    \caption{The proposed method is compared with several state-of-the-art pathological foundation models on EBRAINS (fine-grained and coarse-grained classification), IPD-Brain, BRACS, and TCGA datasets.  Bold numbers indicate superior performance. Subscript \textit{N} denotes \textit{N}-class.}
    \label{tab:res1}
\end{table}

\subsection{Implementation Details}
We implement the proposed method using PyTorch on a single NVIDIA RTX4090 GPU. The AdamW optimizer is employed for training 20K iterations, with an initial learning rate of $1 \times 10^{-5}$, a weight decay of $1 \times 10^{-4}$, and a batch size of 4. WSIs are preprocessed through tissue segmentation, tiling, and feature extraction. Specifically, the Conchv1.5~\cite{ding2024multimodal} is used for feature extraction, while tissue segmentation and tiling are performed using the CLAM toolbox~\cite{lu2021data}. For prompt engineering, we utilize DeepSeek-R1~\cite{guo2025deepseek} to generate discriminative class descriptions, followed by manual verification of the generated text with the assistance of online resources~\cite{luchini2024benign}. During evaluation, balanced accuracy is utilized as the primary classification metric, defined as the mean recall across all classes.

\subsection{Comparison with State-of-the-arts}

We conduct a comprehensive comparison of our proposed FG-PAN framework against several state-of-the-art methods across a suite of histopathology image classification benchmarks.     These include EBRAINS (fine-grained 30-class and coarse-grained 12-class), IPD-Brain, BRACS, and TCGA datasets.     The main experimental results are summarized in Table \ref{tab:res1}.     As shown in the table, the application of our FG-PAN framework consistently enhances the performance of existing baseline models, demonstrating the efficacy and generalizability of its design.

\begin{figure}[t]
    \includegraphics[width=\textwidth]{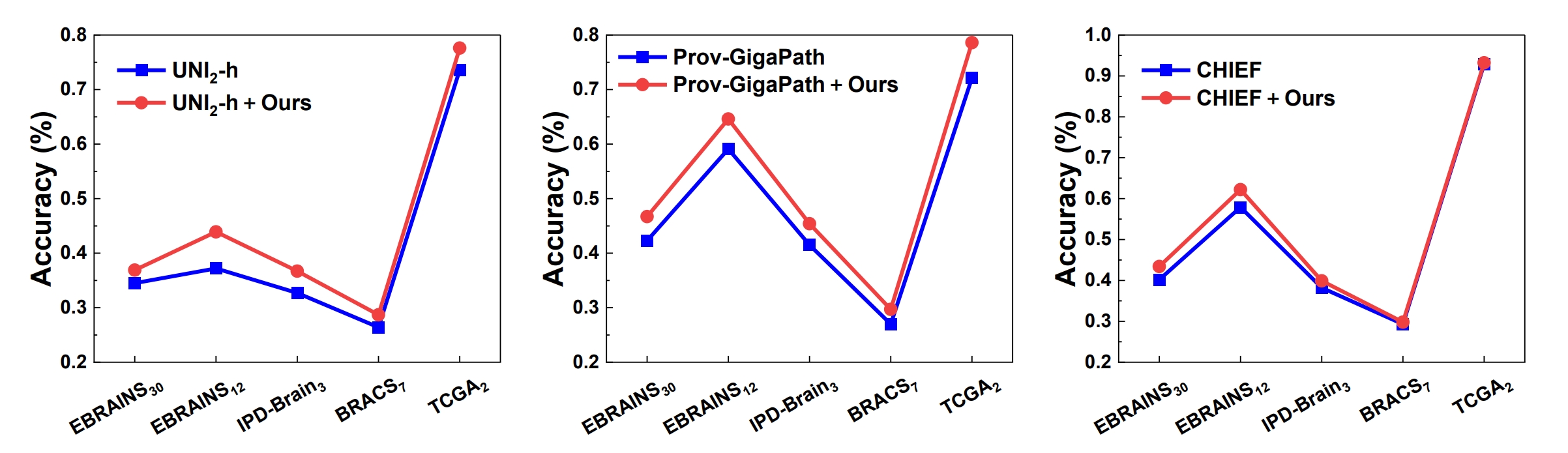}
    \caption{Performance comparison of FG-PAN and baseline methods on three additional pathology foundation models. The line chart shows classification accuracy across different models, demonstrating the consistent superiority of FG-PAN.}
    \label{fig:res3}
\end{figure}

\begin{table}[!t]
\centering

\begin{minipage}{0.45\textwidth}
\centering

\begin{tabular}{l|cc}
\textbf{LLMs}        & Coarse & Fine       \\ \hline
\texttt{[type description]} & 1.910            & - \\\hline
GPT-4o    &        -     & 1.690 \\
DeepSeek-R1    &  -        & 1.660  \\
GPT-4 Turbo    &  -            &    1.670            \\
Gemini 1.5 Pro & - & 1.685 \\
\end{tabular}

\end{minipage}\hfill
\begin{minipage}{0.45\textwidth}
\centering

\begin{tabular}{l|cc}
\textbf{LLMs}        & Coarse & Fine       \\ \hline
Llama-3 70B            & -         & 1.700 \\
Qwen2-72B            &  -               &  1.675           \\
Claude 3 Opus            &   -             &   1.695   \\ 
Grok 3              &       -         &     1.680      \\
\textbf{Average}     &  1.910   &   \textbf{1.682}          \\\hline
\end{tabular}

\end{minipage}

\caption{Performance comparison of the text module using descriptions generated by different LLMs. Results show that the effectiveness of zero-shot classification varies with the source of textual prompts.}
\label{tab:otherllm}
\end{table}

In Table \ref{tab:res1}, we evaluated the impact of FG-PAN on six recent baseline models: UNI~\cite{chen2024towards}, Conch~\cite{lu2024visual}, Conch-v1.5~\cite{ding2024multimodal}, Virchow~\cite{vorontsov2024foundation}, PLIP~\cite{huang2023visual}, and CTransPath~\cite{wang2022transformer}.  For each baseline, we reported both its original performance and the performance after integrating FG-PAN.  The results consistently demonstrate that FG-PAN enhances classification metrics across all datasets and baseline models.  For instance, when applied to Conch-v1.5 on the EBRAINS30 dataset, FG-PAN improved the balanced accuracy from 0.493 to 0.572 and the F1 score from 0.542 to 0.597.  Similarly, on the TCGA2 dataset, FG-PAN elevated the balanced accuracy of the Virchow baseline from 0.735 to 0.772 and improved the F1 score of the PLIP baseline from 0.710 to 0.769.  For the challenging multi-class BRACS7 dataset, FG-PAN consistently increased the F1 scores across all baselines, demonstrating its efficacy in complex classification tasks.


\newcommand{\cmark}{\textcolor{green!70!black}{\ding{51}}}
\newcommand{\xmark}{\textcolor{red}{\ding{55}}}

\begin{table}[!t]
    \centering
    
    \renewcommand\arraystretch{1}
    \setlength{\tabcolsep}{6pt} 
    {
        \small
        \begin{tabular}{@{}ll*{6}{c}@{}}
            \toprule
            \multirow{2}{*}{LWA+GFF} & \multirow{2}{*}{FTDG} & \multicolumn{3}{c}{EBRAINS$_{\text{30}}$} & \multicolumn{3}{c}{EBRAINS$_{\text{12}}$}   \\
            \cmidrule(lr){3-5} \cmidrule(lr){6-8} 
            & & Bacc. & F1 & AUC & Bacc. & F1 & AUC \\ 
            
            \midrule
            
            \multicolumn{1}{c}{\xmark} & \multicolumn{1}{c}{\xmark} & 0.493 & 0.542 & 0.913 & 0.635 & 0.657 & 0.927 \\

            \midrule

            \multicolumn{1}{c}{\xmark} & \multicolumn{1}{c}{\cmark} & 0.539 & 0.552 & 0.918 & 0.663 & 0.704 & 0.925 \\

            \midrule

            \multicolumn{1}{c}{\cmark} & \multicolumn{1}{c}{\xmark} & 0.543 & 0.574 & 0.911 & 0.688 & 0.719 & 0.928 \\

            \midrule

            \multicolumn{1}{c}{\cmark} & \multicolumn{1}{c}{\cmark} & 0.572 & 0.597 & 0.923 & 0.712 & 0.737 & 0.938 \\
            
            \bottomrule
        \end{tabular}
    }
    \caption{Ablation study for the proposed components in our method  on Ebrains dataset.}
    \label{tab:abla1}
\end{table}

\begin{table}[!t]
    \centering
    
    \renewcommand\arraystretch{1}
    \setlength{\tabcolsep}{4pt} 
    {
        \small
        \begin{tabular}{@{}ll*{9}{c}@{}}
            \toprule
            \multirow{2}{*}{LWA+GFF} & \multirow{2}{*}{FTDG} 
            & \multicolumn{3}{c}{IPD-Brain$_{\text{3}}$} 
            & \multicolumn{3}{c}{BRACS$_{\text{7}}$}  
            & \multicolumn{3}{c}{TCGA$_{\text{2}}$} \\
            
            \cmidrule(lr){3-5} \cmidrule(lr){6-8} \cmidrule{9-11} 
            & & Bacc. & F1 & AUC & Bacc. & F1 & AUC & Bacc. & F1 & AUC \\ 
            
            \midrule
            
            \multicolumn{1}{c}{\xmark} & \multicolumn{1}{c}{\xmark} & 0.486 & 0.412 & 0.791 & 0.289 & 0.291 & 0.786 & 0.679 & 0.712 & 0.772 \\

            \midrule

            \multicolumn{1}{c}{\xmark} & \multicolumn{1}{c}{\cmark} & 0.506 & 0.491 & 0.816 & 0.295 & 0.305 & 0.794 & 0.745 & 0.774 & 0.853 \\

            \midrule

            \multicolumn{1}{c}{\cmark} & \multicolumn{1}{c}{\xmark} & 0.514 & 0.515 & 0.808 & 0.313 & 0.306 & 0.793 & 0.765 & 0.783 & 0.861 \\

            \midrule

            \multicolumn{1}{c}{\cmark} & \multicolumn{1}{c}{\cmark} & 0.522 & 0.530 & 0.833 & 0.329 & 0.334 & 0.796 & 0.797 & 0.801 & 0.891 \\
            
            \bottomrule
        \end{tabular}
    }
    \caption{Ablation study for the proposed components in our method  on IPD-Brain, BRACS and TCGA dataset.}
    \label{tab:abla2}
\end{table}


\begin{figure}[t]
    \includegraphics[width=\textwidth]{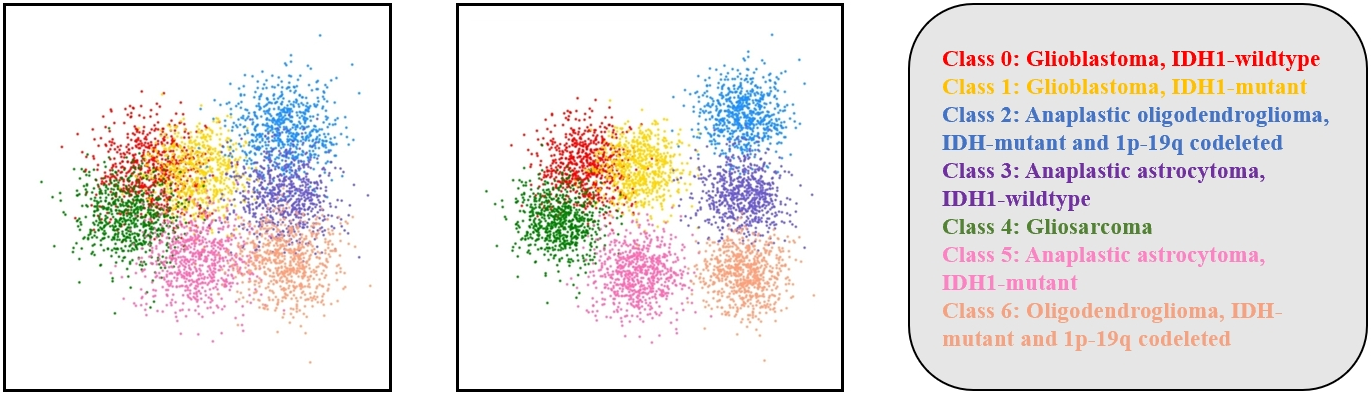}
    \caption{\textbf{t-SNE visualization} of feature distributions using original versus LLM-generated fine-grained textual descriptions. The comparison highlights improved class separability and semantic alignment achieved by the LLM-enhanced prompts.}
    \label{fig:tsne}
\end{figure}

In addition to the models presented in the tables, we conducted a further evaluation on three other prominent foundation models: UNI2-h~\cite{chen2024towards}, Prov-GigaPath~\cite{xu2024whole}, and CHIEF~\cite{wang2024pathology}, with the results visualized in Figure \ref{fig:res3}.   For these models, we followed the same evaluation protocol, applying our FG-PAN framework to assess the performance gains.

Although these models represent diverse architectural approaches, our method again yields substantial performance enhancements.     A notable example is on the TCGA$_2$ dataset, where our framework elevates the balanced accuracy of the Virchow baseline from 0.735 to 0.772 and the F1 score of the PLIP baseline from 0.710 to 0.769.     Furthermore, on the challenging multi-class BRACS$_7$ dataset, our method consistently improves the F1 scores for all three baselines, which demonstrates its strong capabilities in complex classification tasks and validates the superiority of our proposed framework.

\subsection{Ablation Study}

In this section, we systematically evaluate the effectiveness of each key module in the FG-PAN method through a series of ablation experiments. Additionally, we analyze the impact of text descriptions generated by different LLMs on the distribution of the feature space.

\textbf{Ablation Study of Key Modules.} To validate the effectiveness of each key module in FG-PAN, we conducted systematic ablation experiments on multiple datasets, as shown in Table \ref{tab:abla1} and Table \ref{tab:abla2}. The experimental results demonstrate that introducing either the Fine-grained Text Description Generation module or the Local Window Attention and Gated Feature Fusion module individually leads to significant performance improvements. For instance, on the EBRAINS$_{30}$ dataset, both balanced accuracy and Weighted-F1 score show notable increases. When both modules are combined, the model achieves optimal performance, indicating that spatial modeling and fine-grained semantic description play complementary roles in pathological image understanding. This trend is consistently observed across other datasets, further confirming the rationality and strong generalization capability of the FG-PAN approach.

\textbf{Discriminative Power of Text Descriptions Generated by Different LLMs.} In addition to the ablation of structural modules, this study systematically analyzes the impact of fine-grained text descriptions generated by different LLMs on classification performance, as shown in Table \ref{tab:otherllm}. The results indicate that when fine-grained pathological descriptions generated by mainstream LLMs such as GPT-4o~\cite{hurst2024gpt}, DeepSeek-R1~\cite{guo2025deepseek}, Gemini 1.5 Pro~\cite{team2024gemini}, and Llama-3 70B~\cite{dubey2024llama} are used, the average inter-class feature space distances consistently range between 1.66 and 1.70. This performance is stable and comparable across models, and is significantly better than using only category names, which yields a distance of 1.910. These findings suggest that text descriptions generated by different LLMs can effectively increase the feature separation between categories, providing the model with more discriminative semantic information and laying a solid foundation for zero-shot classification.

\textbf{t-SNE Visualization Analysis.} To further intuitively demonstrate the impact of fine-grained text descriptions on the distribution of the feature space, we visualized the feature distributions using t-SNE, as shown in Figure \ref{fig:tsne}. The results show that when only category names are used, the feature distributions of different classes exhibit significant overlap, with blurred boundaries between categories. In contrast, after introducing fine-grained text descriptions, the distributions of each class in the feature space are clearly separated, with more compact intra-class features and more distinct inter-class separation. This phenomenon visually confirms the crucial role of the fine-grained text module in enhancing the model’s discriminative ability and the separability of the feature space.

\section{Conclusion}

In this paper, we propose FG-PAN, a novel framework for fine-grained zero-shot classification of brain tumor subtypes based on whole slide images. By integrating local window attention and gated feature fusion, our method effectively models spatial dependencies and enhances the discriminability of patch-level features. Furthermore, the fine-grained text description generation module leverages large language models to construct semantically rich and pathology-aware class prompts, significantly improving the separability of different categories in the feature space. Through extensive experiments and ablation studies on multiple digital pathology benchmarks, we demonstrate that each component of FG-PAN contributes to performance gains, and their combination leads to state-of-the-art results across diverse datasets. Notably, our approach exhibits strong generalization and robustness, maintaining high accuracy even when applied to out-of-domain cohorts and using different LLMs for prompt generation. These results validate the effectiveness and versatility of FG-PAN in addressing the challenges of fine-grained zero-shot classification in computational pathology. In the future, we plan to further explore the integration of multimodal clinical data and advanced prompt engineering techniques to enhance the interpretability and clinical applicability of our framework.

\bibliographystyle{elsarticle-harv}
\bibliography{paper}

\end{document}